**Human-like generalization in a machine through predicate learning**

Leonidas A. A. Doumas[1*], Guillermo Puebla[1], Andrea E. Martin[2]

[1]School of Philosophy, Psychology, and Language Sciences, The University of Edinburgh
[2]Max Planck Institute for Psycholinguistics, Nijmegen, The Netherlands
*Address for correspondence: 7 George Square, Edinburgh, EH8 9JZ, UK; email: alex.doumas@ed.ac.uk

**Abstract:**
Humans readily generalize, applying prior knowledge to novel situations and stimuli. Advances in machine learning and artificial intelligence have begun to approximate and even surpass human performance, but machine systems famously struggle to generalize information to untrained situations. We present a model that demonstrates human-level extrapolatory generalization and does so by exploiting well-established principles from neuroscience. This model is trained to play one video game (Breakout) and performs one-shot generalization to a new game (Pong) with different rules, dimensions, and characteristics. The model generalizes because it learns structured representations that are functionally symbolic (viz., a role-filler binding calculus) from unstructured training data. It does so without feedback, and without requiring that structured representations are specified a priori. Specifically, the model uses neural co-activation to discover which characteristics of the input are invariant and to learn relational predicates, and oscillatory regularities in network firing to bind predicates to arguments. We argue that models of human cognition must account for far-reaching and flexible generalization, and that in order to do so, they must be able to discover symbolic representations from unstructured data, and ideally do so by capitalizing on basic principles from neurophysiology. Only then can models begin to adequately explain where human-like representations come from, why human cognition is the way it is, and why it continues to differ from machine intelligence in crucial ways.

**Summary:** We present a machine system that learns predicates from experience, without feedback or pre-specified knowledge; the system uses neural oscillations that naturally occur in the network to compute. After learning, the system uses the predicates it learned from unstructured data to support one-shot generalization from one video game to another.

There is vibrant controversy about the relationship between artificial intelligence systems, the human mind, and the principles of neural computation (e.g., *1-5*). For example, children can easily build Lego constructions in either a horizontal or vertical direction once they know how to attach the pieces together, yet a machine learning system would struggle to transfer its knowledge even in such a basic way[1].

This disconnect in behavior between child and machine highlights a core aspect of the human mind: the ability to generalize to situations 'outside the training set'. In domains as various as sports, music, language, mathematics, or gaming, people learn from their environment, refine what they know from experience, and employ strategies that they learned in one context to solve problems in a new context. Despite its ubiquity in human cognition, however, the ability to generalize across contexts presents a striking lacuna for machine learning systems.

Here we present a machine system that generalizes like the child playing with Legos. To demonstrate this point, we solve a classic generalization problem in machine learning (*5*): the transfer of knowledge and play behavior learned from one video game (Breakout, where a paddle hits a ball by moving along the horizontal axis and points are scored by hitting blocks), to a different game, (Pong, where two paddles hit a ball along the vertical axis and points are scored by hitting the ball past the opposite paddle). Being able to play Pong once you know how to play Breakout mimics the generalization behavior of humans and, importantly, contrasts starkly with current state-of-the-art machine systems. To our knowledge, no other machine system can play a new, untrained video game based only on representations learned from another game.

Importantly, our system achieves generalization or task transfer outside the training set by learning functionally symbolic representations (viz., a role-filler binding calculus) from experience, in an unsupervised fashion (without feedback), and without requiring pre-specified structured (i.e., compositional or symbolic) knowledge. We call this process *predicate learning*. We show that the resulting representations allow the model to achieve one-shot generalization between the video games Breakout and Pong (as humans routinely do, and machine systems famously struggle with). Crucially, our system exploits model-internal 'neural' oscillations in order to achieve predicate learning in a manner that is similar to how human cortical networks and neuronal assemblies communicate with and modulate each another (*6*).

Recent advances in machine learning (e.g., *7-9*) have produced deep neural networks (DNNs) that reach and even exceed human levels of performance on a range of cognitive tasks. For example, recent DNNs learned to master an impressive number of Atari games (*9*). DNNs are general, in that they can learn to perform a variety of tasks without *a priori* background knowledge. Nevertheless, while DNNs readily perform *interpolation* (i.e., generalization to untrained items from within the bounds of the training set, or which have the same statistical properties as the training data), they struggle to perform *extrapolation* (i.e., generalization to items from outside the bounds of the training set). While a human learner who plays one video game—like Breakout—will quickly generalize to playing a new game—like Pong—DNNs trained to play one game must be retrained in order to play any additional games (*9*).

That DNNs have difficulty extrapolating their knowledge is well known and appears to stem from their explicit lack of structured representations (*10-13*). By contrast, accounts of how humans generalize are frequently based on powerful symbolic languages or graph grammars that include structured relations (predicates), which can be promiscuously applied to new arguments (*14-17*). Structured representations allow the flexible transfer of information across contexts because the same representations can be used to characterize very different situations (Fig. 1). For example, one recent model of hand-writing recognition (*18*) can learn to recognize and even recreate novel hand-written characters after only a single exposure, because it represents those characters in terms of predicates, *attached-at-start*(*x,y*), *attached-along*(*x,y*), and *attached-at-end*(*x,y*), that relate possible line segments together. However, models that exploit structured representations face a challenge that is complementary to that faced by DNNs: They require the modeler to specify the vital structured representations in advance of any learning (*14-19*). So, while structure-based models generalize more flexibly than DNNs, they do not possess the same capabilities to learn from limited starting states.

---

[1] This example comes from Gary Marcus in the article http://www.sciencemag.org/news/2018/05/how-researchers-are-teaching-ai-learn-child.

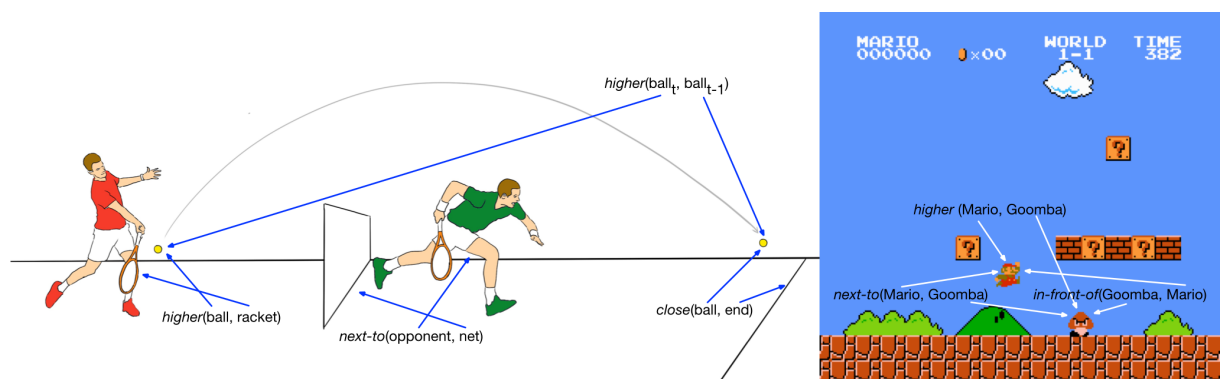

**Fig. 1.**
Representations of relations and predicates in two disparate domains: video games and tennis. Many of the same relations that support reasoning about a strategic shot in tennis allow also support reasoning about a strategic move in a game of Super Mario Bros. (e.g., the *higher* relation comes into play when planning a drop shot in tennis, or a jump attack by Mario).

We describe a neural network model that bridges the gap between these frameworks. The model is a general learner that discovers powerful symbolic representations from experience. The model is trained via *predicate learning*, and the representations that it learns allow it to reach human-level performance on some tasks, and, crucially, to generalize and transfer that expertise in a human-like fashion, quickly (in one shot) and without explicit feedback. The architecture, called DORA (*Discovery of Relations by Analogy*; *20*), is based on two influential ideas from cognitive science and neuroscience. First, learning and generalization depend upon a process of comparison (e.g., *21-23*). Second, information in neural computing systems can be carried by the oscillatory regularities that emerge as its component units fire (e.g., *6, 24-27*). In brief, DORA uses unsupervised comparison to discover which characteristics of the input are invariant, and to learn functional predicates; it then applies these predicates to arguments in a symbolic fashion, using oscillatory regularities to dynamically bind predicates and arguments. DORA learns representations that are functionally and formally symbolic from training data, without feedback, and without requiring that structured representations be specified a priori. We demonstrate that the capacity for predicate learning produces representations that, coupled with "off-the-shelf" preprocessing and reinforcement learning tools, support mastery of a video game, and, more importantly, that the representations then support transfer of expertise in that game to a new game in one shot, extrapolating knowledge in a manner that closely resembles human players.

**Predicate learning**

Our theory of predicate learning is instantiated in a neural network model called DORA (*Discovery of Relations by Analogy*; *20*), which is descended from of the symbolic-connectionist system LISA (*Learning and Inference with Schemas and Analogies*; *15,16*). DORA learns to represent structured (i.e., functionally symbolic) representations from unstructured examples without feedback. DORA is based on traditional neurocomputing principles but differs in important ways from conventional neural networks. Below we present an overview of the DORA model and how it performs predicate learning. Details of the operations described below can be found in *20*, with the exception of the local energy operations that are described in detail in the Methods section. Full details of the entire model also appear in the Supplemental Methods. Code for the model is available online (see Acknowledgements).

Table 1 summarizes the architectural assumptions of the model. The basic network macrostructure is presented in Fig. 2. DORA consists of a long-term-memory (LTM) composed of bidirectionally connected layers of nodes. We refer to nodes in LTM as token units (or tokens). Units in the lowest layer of LTM are connected to a pool of feature units. Token units are yoked to integrative inhibitors that integrate input from their yoked unit and active token units in higher layers, and fire after reaching a threshold. The yoked inhibitors serve the purpose of implementing phasic firing and refractory periods in the token units (which are important for implementing dynamic binding in the network; described below).

Table 1. Architectural assumptions of the DORA model.

| Architectural Structure | Function |
|---|---|
| Layered hierarchical network of bidirectionally connected units (token units) comprises a long-term memory (LTM). | Allows the network to represent information in a hierarchy of conjunctive links. |
| The lowest layer of token units is connected to a pool of feature units. | Units that encode semantic features of a stimulus in the environment, or which were activated in elsewhere in the network |
| Token units have falling phases in firing (implemented via integrative inhibitors yoked to each token unit). | Allows oscillations and time sharing that underlie dynamic binding. |
| Floating working memory banks within LTM, which correspond to potentiated units (or focus of attention (FOA) and active memory (AM) in cognitive frameworks). | Correspond to potentiated units, or to the focus of attention (FOA) and active memory (AM) in cognitive frameworks. Allow for comparison between independent sets of units. |
| Token units are laterally inhibitive (units in the same layer inhibit one another) within but not across floating banks. | Allows units in AM to compete to respond to patterns of activations on the feature units. Important for time-based binding. |
| Activation flows from FOA to other token units via activated feature units. | Allows comparison across banks. Avoids the problems feed-forward architectures face where input and output are to definitionally independent spaces. |
| Excitatory mapping connections learned between within layer co-active units in FOA and AM via modified Hebbian algorithm. | Allows the system to learn correspondences between firing patterns in FOA and AM. |

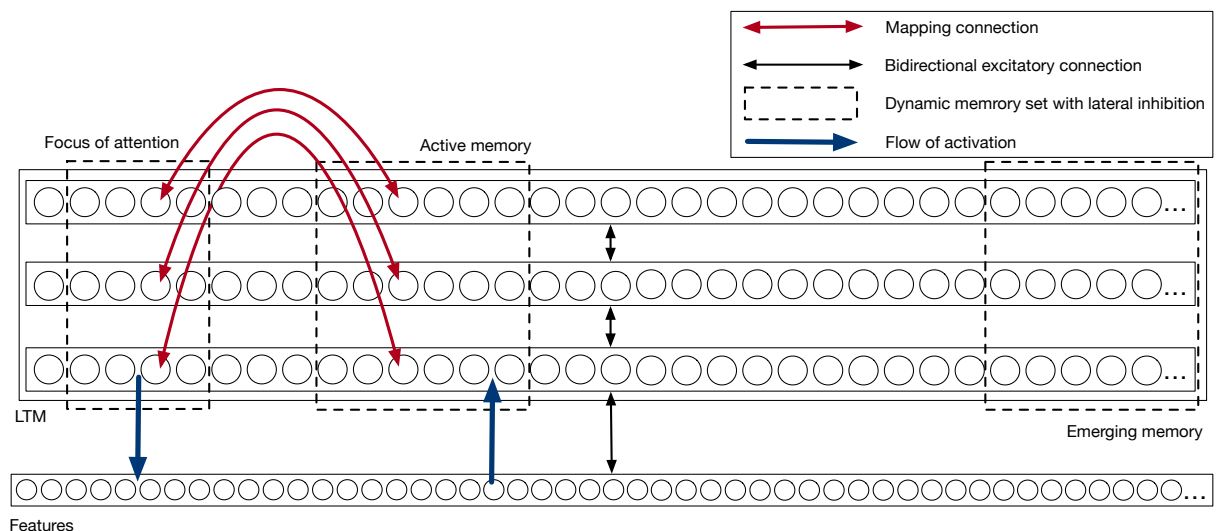

**Fig. 2.**
Macrostructure of the DORA network. LTM (long-term memory) is composed of layers of bidirectionally connected units (called tokens) yoked to integrative inhibitors (each circle represents a token unit and its yoked inhibitor). Dashed boxes represent floating memory sets, potentiated units that are laterally inhibitive corresponding to DORA's focus of attention (what the model is currently processing), and active memory (or working memory in cognitive terms).

Floating memory sets, groups of potentiated units, correspond to attention/working memory (WM) within a cognitive framework. One floating memory set corresponds to DORA's current focus of attention (FOA), one to DORA's active memory (AM), and one to emerging memory (EM). Token units are laterally inhibitive (units in the same layer inhibit one another) within, but not across, sets. Activation in the model flows from the FOA to token units in the AM and LTM via the shared pool of feature units. Excitatory connections, called mapping connections, are learned within-layer between co-active units in FOA and AM via a modified Hebbian algorithm (*16,20*).

DORA is a settling network and thus does not have inputs or outputs in the conventional sense. Rather, DORA has states that produce new states. DORA starts in some state (e.g., a set of units in FOA) and settles (e.g., with some units active in FOA and AM). Due to the refraction of nodes, this settling state will at some point become upset, and DORA will again settle, producing a new state (e.g., a different set of active units in FOA and AM), and so on. In each settled state, DORA attempts to find mapping between active units in FOA and AM, or to learn (as predicate learning or as generalization; see below).

DORA's functional processing is summarized in Fig. 3. All functional operations described in Fig. 3 are accomplished using traditional neurocomputing principles (*20* and supplemental methods). DORA starts with a set of potentiated units in FOA (chosen at random from LTM, or based on DORA's current perceptual state; e.g., a video-game screen shot). As DORA runs, units in FOA—as a consequence of lateral inhibition and yoked inhibitors—naturally oscillate. Activation flows from FOA to features and then to token units in the rest of LTM. Active units in LTM are retrieved into AM (become potentiated) using a Luce (*28*) choice rule. DORA learns mapping connections (positive weighted connections) between co-active units in FOA and AM. After mapping, DORA learns predicates (first single-place predicates, and eventually multi-place predicates) via unsupervised predicate learning. As a consequence of the temporal dynamics produced in the network, these predicates can be dynamically bound to arguments (see below).[2] Finally, the current instantiation of DORA exploits an energy signal that occurs naturally in distributed computing systems to discover an invariant signal for, and to learn a circuit to respond to, instances of similarity and relative magnitude (see below and Methods for details). As a result, the predicates that the network learns instantiate representations of relative magnitude.

---

[2] DORA learns representations that are formally equivalent to a role-filler binding calculus (Doumas & Hummel, 2005; 2012). In a role-filler binding calculus, all predicates are single-place, and multi-place relations are represented as linked sets of single-place predicates. For example, a multi-place concept like *above*(*x*,*y*) is represented as *higher*(*x*)+*lower*(*y*), where the agent and patient roles of the *above* relation are represented explicitly and linked via the '+' operator. A role-filler binding system is formally equivalent in expressive power to a traditional predicate system.

| Processing step | | Description |
|---|---|---|
| Items $F$ in FOA | < | Potentiated units in FOA. |
| $w(\tau_i, \tau_j, \sigma_n) \leftarrow LE(\tau_i, \tau_j)$ | < | Update weights in the local entropy circuit. Full details in Methods. |
| **For** $i \in F$ : | < | $F$ is the units at the highest token layer for any set of connected units. |
|    **Until** $Y_i > \theta_i$ : | < | While unit $i$'s yoked inhibitor has not fired. |
|       $a_i = 1$ | < | Activation of unit $i$ clamped to 1. |
|       $\overleftarrow{n_j}$ | < | Inputs to all network units $j$ updated. |
|       $a_j \leftarrow f(n_j)$ | < | Activations and states of all network units $j$ updated. |
|       $\overleftarrow{Y_k}$ | < | Input to yoked inhibitors updated. |
|       **If** $Y_k > \theta_k$ : | < | If the local inhibitor is above threshold. |
|          $\overrightarrow{I_L}$ | < | Inhibitor fires (refresh all units at layer $L$ of unit $k$ and below). |
|       **End If** | | |
|       **When** retrieving : | < | When the network is retrieving from LTM. |
|          $E(G) \leftarrow f(A(G), A(H)), H \neq G$ | < | Potentiated units into AM via the Luce choice rule. |
|       **End When** | | |
|       **When** mapping : | < | During mapping. |
|          $\Delta m_{jk} \leftarrow M(a_j, a_k)$ | < | Mapping connections $M$ between FOA and AM updated (20). |
|       **End When** | | |
|       **When** predicate learning : | < | During predicate learning (20). |
|          **If** $T(m_{FOA,AM})$ : | < | If there are mapping connections between units in FOA and AM. |
|             $A(\tau_{j,L=1}) \leftarrow \sim A(\tau_{x,L>1})$ | < | Token unit in AM layer $L=1$ is activated, $A$, when no tokens at higher layers are active. |
|             $A(\tau_{j,L}) \leftarrow A(\tau_{x,L-1})$ | < | Token unit in AM layer $L$, above a layer $L-1$ with active token units, is activated. |
|             $\Delta w_{jk} \leftarrow H(a_j, a_k)$ | < | Weights updated via Hebbian learning. |
|          **End If** | | |
|       **End When** | | |
|       **When** refining : | < | During refinement (schema induction; 20). |
|          **If** $T(m_{FOA,AM})$ : | < | If there are mapping connections between units in FOA and AM. |
|             $A(\tau_{j,EM}) \leftarrow M(\tau_{x,FOA}, \tau_{y,AM})$ | < | Token unit $j$ in EM is activated matching active FOA unit with mapping connections. |
|             $\Delta w_{jk} \leftarrow H(a_j, a_k)$ | < | Weights updated via Hebbian learning. |
|       **End When** | | |
|       **When** generalizing : | < | During generalization (20). |
|          **If** $T(m_{FOA,AM})$ : | < | If mapping connections between units in FOA and AM. |
|             $A(\tau_{j,AM}) \leftarrow \sim M(\tau_{x,FOA})$ | < | Token unit $j$ in AM is activated matching active FOA units with no mapping connections. |
|             $w_{jk} \leftarrow H(a_j, a_k)$ | < | Weights updated via Hebbian learning. |
|          **End If** | | |
|       **End When** | | |
|    **End Until** | | |
|    **If** $Num(w_{jk}) < 2$ : | < | If token units $j$ is not multiply connected. |
|       $zero(w_{jk})$ | < | Zero weights from token $j$. |
|    **End If** | | |
| **End For** | | |

**Fig. 3.**
Functional description of processing in DORA.

Applied iteratively, these processes produce structured predicate representations from unstructured inputs and support relational generalization (extending knowledge from one system to reason about another; see *20*), both without supervision. These computations have allowed the model to account for over 50 phenomena from the literature on human cognitive development (*20, 29*-34), relational reasoning (*20, 35-38*), and language processing (*39, 40*).

To illustrate the model's operation, we describe the stages of the model as it learns predicates and relations and performs binding. These outcomes result from the architecture described directly above.

DORA starts with representations of objects coded as flat vectors (e.g., of features). For the current simulations, we use a very basic visual preprocessor that performs edge detection (via contrast) to identify and encode simple objects. The preprocessor had an inbuilt bias: any enclosed edges were treated as a single object. The preprocessor returns an encoding of each object as in the form of retinal impingement and colour. This information roughly corresponds to the total retinal area of the object and the enervation of the superior, inferior, lateral, and medial rectus muscles in reaching the (rough) center of the object from a reference point (*41*) (this information is currently encoded as the raw pixels and direction (specific muscle) between the rough object center and the reference point; we use the edge of the image, although results are the same using a central reference point), and the RGB encoding of the pixels composing the object. One consequence of the particular encoding is that the model shows the same bias for the cardinal directions observed in humans (*42*). The preprocessor's operations are all based on information available in the early visual system (e.g., the visual system detects edges by contrast (*43*) represents extent information in the retinal image in absolute terms (*41, 44-46*), but it remains a massive oversimplification of human vision.

DORA isolates and explicitly represents features or properties of objects by exploiting the natural intersection highlighting that occurs when two distributed representations are compared (Fig. 4A). Comparing (and co-activating) distributed representations of items, $O_i$ and $O_j$, naturally reveals shared and unshared features of the two items. Specifically, features shared by $O_i$ and $O_j$ will receive roughly twice as much input and, therefore, become roughly twice as active as unshared features (Fig. 4A,v). DORA exploits this emergent regularity by learning connections between a recruited unit, $r_1$, and active feature units by Hebbian learning (Fig. 4A,vii). The result is a unit explicitly encoding the intersection of the compared items, and a solution to the isolation problem. Concurrently, a recruited unit at a higher layer of the network learns connections between the active $r_1$ and $O_j$ units (Fig. 4A,viii), conjunctively coding a link between them.

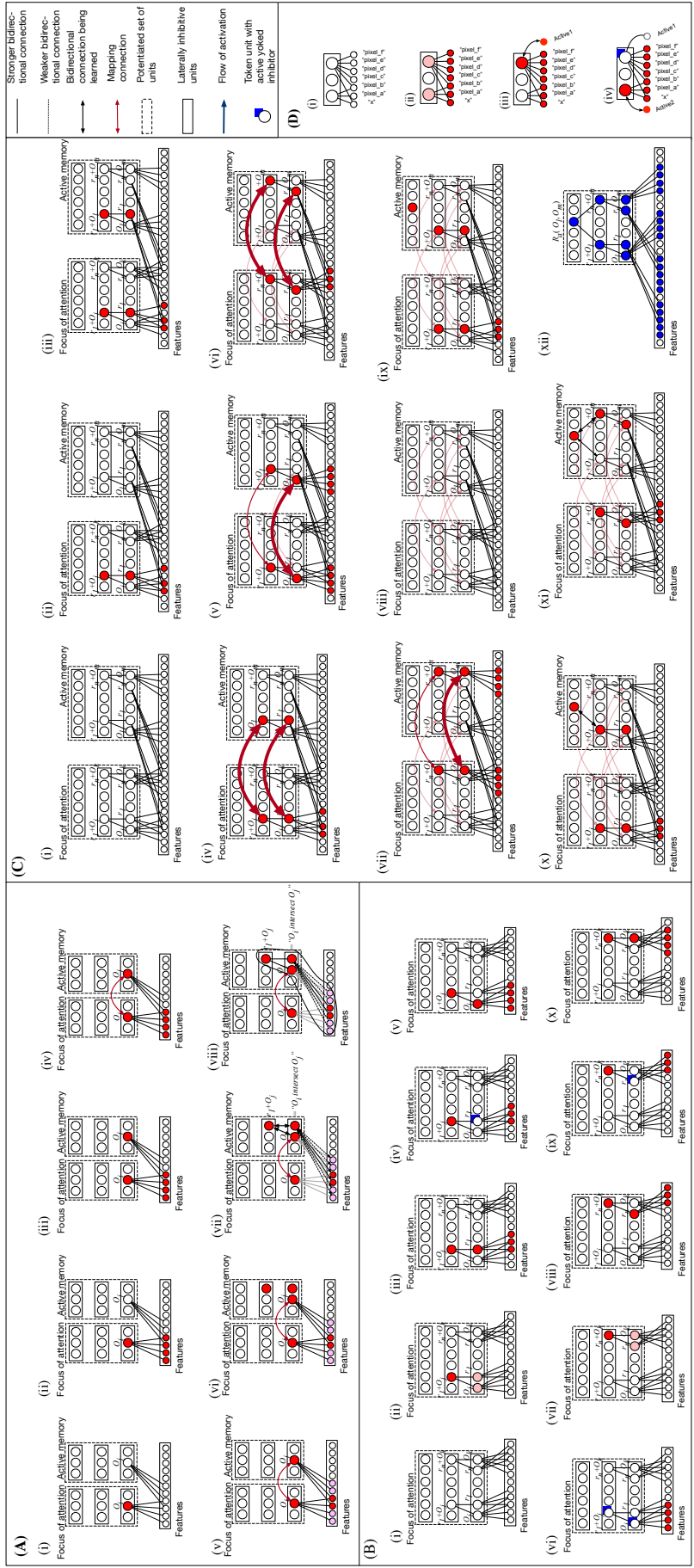

**Fig. 4.**
Learning in DORA. (A) Isolation and explicit representation of invariant properties via comparison. (i) Two objects, $O_i$ and $O_j$, are in FOA and AM (respectively), and $O_i$ becomes active. (ii) $O_i$ passes activation to its constituent features. (iii) Oj becomes active. (iv) $O_i$ and $O_j$ are mapped. (v) $O_i$ and $O_j$ are coactive, and so any shared features receive roughly twice as much input and become roughly twice as active as unshared features. (vi) A unit $r1$ and a token unit at the layer above the active $O_j$ unit are recruited and activated, and (vii) learn connections to the active units in proportion to their activation (via Hebbian learning). (viii) Unit $r1$ explicitly codes for the intersection of $O_i$ and $O_j$. (B) Learned representations are dynamically bound to arguments by systematic phase-lag of firing. (i) The pair $r1+O_j$ learned in (A), and another pair, $rn+O_k$ learned as in (B). (ii) Unit coding for $r1+O_j$ conjunction becomes active, and passes activation to the $r1$ and $O_j$ units, which compete, via lateral inhibition, to become active. (iii) $r1$ becomes active first. (iv) $O_j$ become active next. The direct sequence of firing marks the binding of $r1$ to $O_j$. (v-vii) A similar pattern emerges when the $rn+O_k$ role-binding becomes active. (C) Lower arity representations are conjoined into multi-place relational structures. (i) Role-filler pairs $r1+O_j$ and $rn+O_k$ from (B) are compared to role-filler pairs $r1+O_l$ and $rn+O_m$ (learned by the network at some other timepoint). (ii) As a result of phase-lag binding in (C), bound role-filler pairs will oscillate. Units coding for $r1$ in the focus of attention become active. (iii) Activation flows from the focus of attention to active memory via the shared feature units activating units coding for $r1$ in active memory. (iv) The network maps (bold red arrows) co-active units in focus of attention and active memory. (v) The units coding for $O_i$ and $O_j$ similarly become active, and the maps them. (vi) Units coding for $rn$ then become active, and the network maps them. (vii) Units coding for the arguments of $rn$ are similarly mapped. (viii) The network has learned mappings between corresponding (structurally similar) units in focus of attention and active memory. (ix) As a result of the predicate learning algorithm, a unit a higher layer of the network is recruited as the units in the focus of attention and active memory become active again. (x) The recruited unit learns connections to role-filler pairs as they become active, first the $r1+O_j$ pair, (xi) then the $rn+O_l$ pair. (xii) The result is a structure (nodes colored blue) that functions as a multi-place predicate. (D) Invariant responses to similarity and relative magnitude emerge when objects with absolute magnitudes are compared. (i) Two representations of objects (large circles) with different values on some dimension are compared. (ii) Magnitude representations are activated, and the objects compete, via lateral inhibition, to become active (red = more active unit). (iii). The object with the greater magnitude wins the competition—the invariant signal for greater magnitude—and learns a connection to a node indicating the competition winner. (iv) The object with the lesser magnitude becomes active next—the invariant signal for lesser magnitude—and learns a connection to a node indicating the competition loser.

DORA dynamically binds the representations learned via comparison to arguments to produce functional single-place predicates. When laterally inhibitive units are linked by a conjunctive node, they will naturally oscillate out of synchrony and in direct sequence (phase-lag-1) when the conjunctive unit becomes active (Fig. 4B,i-v). DORA exploits this emergent oscillatory pattern as a binding signal. Specifically, bound units fire in direct sequence, and systematically out of synchrony with other bound pairs (Fig. 4B,vi-ix). In short, identity information is carried by *what* units are active, and binding information is carried by *when* those units are active. This binding signal is dynamic in that the same units are used to represent complementary bindings through different firing orders (e.g., binding $r_1$ to object $O_j$ and $r_2$ to object $O_i$, amounts to the units coding for $r_1$ firing, followed by the units for $O_j$, and then the units coding for $r_2$ firing, followed by the units coding for $O_i$). As we demonstrate below the resulting representations are functional predicate-object, or role-filler, pairs.

DORA solves the problem of learning higher-arity (i.e., multi-place) representations via comparison. As noted above, DORA learns representations in a role-filler format (*47,48*). That is, DORA learns a representation of a multi-place relation by linking single-place predicate representations of the constituent roles into a single structure (*20*). When structurally similar sets of bound elements are in FOA and AM (as in Fig. 4C,i), DORA will settle in states where corresponding units in FOA and AM are active together and are mapped (Fig. 4C,ii-vii). In short, as a consequence of mapping and dynamic binding corresponding units across FOA and AM will oscillate systematically (Fig. 4C,ii-vii). This pattern is a reliable signal that DORA exploits to learn multi-place relational structures. DORA recruits a unit in a higher layer of the network (Fig. 4C,ix), and learns connections between that unit and active conjunctive token units in the layer below by Hebbian learning (Fig. 4C,viii-xi).

The resulting structure (Fig. 4C,xii) is a functionally symbolic multi-place relational representation. At the layer of feature units, objects, properties, and roles are coded in a distributed fashion. At the next later, collections of properties are conjunctively coded as tokens of individual objects and predicates. At the next layer, role-filler pairs are conjunctively bound. At the top layer, conjunctively coded role-filler pairs are themselves linked into whole multi-place relations. The resulting structure allows for storage in LTM. When the representation becomes active, dynamic binding information is carried in the temporal patterns of firing that emerge, with tokens for individual predicates and objects bound by their firing phase-lag (Fig. 4B).

Finally, the predicate representations that DORA learns represent relative relational concepts (e.g., "same", "more", and "less") as DORA exploits the property that absolute magnitude information for spatial dimensions is encoded in an analog fashion by neural systems (*28-30*). When two objects, $O_i$ and $O_j$, that differ in value on some magnitude dimension (e.g., size) are co-activated (Fig. 4D,i), and the units coding $O_i$ and $O_j$ compete to stay active (via lateral inhibition; Fig. 4D,ii). The object with the larger dimensional value will win the completion and become active first (as it is coded by a greater extent of units; Fig. 4D,iii), while the lesser object will lose and become active next (Fig. 4D,iv). By contrast, when two items of the same magnitude are compared, they will settle into a state of mutual co-activation. DORA exploits these invariant emergent patterns by learning specific responses to them. DORA learns to activate a unit (or set of units) in response to winning the competition (the invariant for "moreness"), another unit (or set of units) in response to losing the competition (the invariant for "lessness"), and another unit (or set of units) in response to mutually co-activation (the invariant for "sameness"), and learns connections between these units and any active object units via Hebbian learning (Fig. 3A,iii-iv). Details of learning this circuit in Methods.

**Results**

Our aim was to test generalization outside the training space. We compared two broad classes of implementations: one with predicate learning, and several without. We situated DORA's predicate learning algorithm between a visual preprocessor (see above), and tabular Q-learning (*49, 50*) (see Fig. 5) in order to test whether predicate learning supports learning representations that allow a system to generalize outside the training space (i.e., to extrapolate).

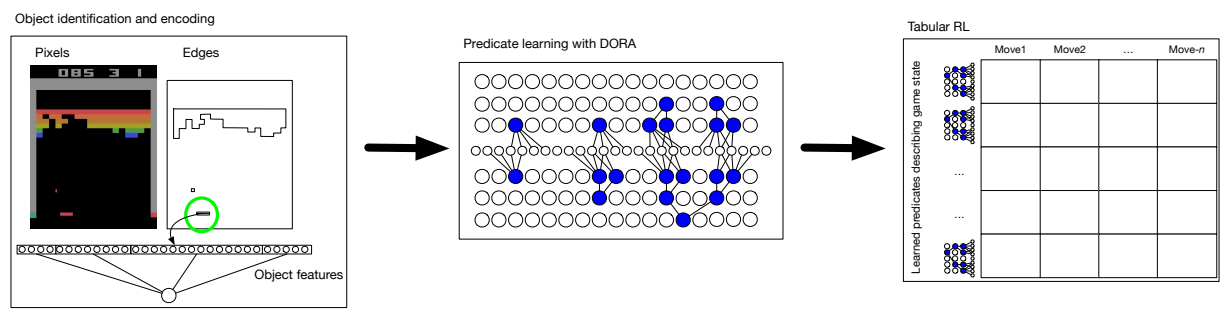

**Fig. 5.**
Predicate learning in context. The DORA model is situated within a visual preprocessor, and a tabular Q-learning mechanism.

We compared (1) an implementation of DORA with Q-learning against (2) DQN; (3) DQN with preprocessed inputs; (4) a supervised (i.e., back-prop) deep neural network (DNN) with preprocessed inputs with fixed frame skipping, and (5) a supervised DNN with preprocessed inputs with fixed frame skipping; (6) Humans (two Breakout and Pong novices). The other networks served as control conditions to test whether predicate learning can underlie extrapolatory generalization, and whether extrapolatory generalization can be achieved without predicate learning. In addition, we also ran simulations using DORA with Q-learning using random frame skipping as well as with supervised learning (see supplemental results).

Specifically, we trained all these systems to play one videogame (Breakout), and then tested their ability to generalize to a different videogame (Pong) without any explicit training. Finally, we evaluated these systems' ability to switch back to playing the original game, after time spent learning to play the second.[3] See Methods for details of all simulations.

For the first 250 games of Breakout, DORA made random moves, generating game states from which it learned structured representations in an unsupervised manner as described above. DORA successfully learned predicate representations encoding to instances such as *moreY* (object1, object2) (effectively a representation of the relation *higher*), and *moreX* (object1, object2) (effectively a representation of the relation *right-of*)[4]. As demonstrated in the supplemental results, the representations it learned are functionally symbolic.

DORA then attempted to learn to play Breakout using the representations that it had learned during the first 250 games to represent the current game screen and then made a response. Associations between these learned representations and successful moves were learned via tabular Q-learning (e.g., *49, 50*). While much more sophisticated methods for representational selection exist, we employed this very simple solution as a proof of concept.

Fig. 6A shows the performance of all networks on Breakout as an average score of the last 100 games played (after reaching plateauing performance), and a high score. Unsurprisingly, all systems performed quite well, reaching levels of performance that matched or exceeded human participants.

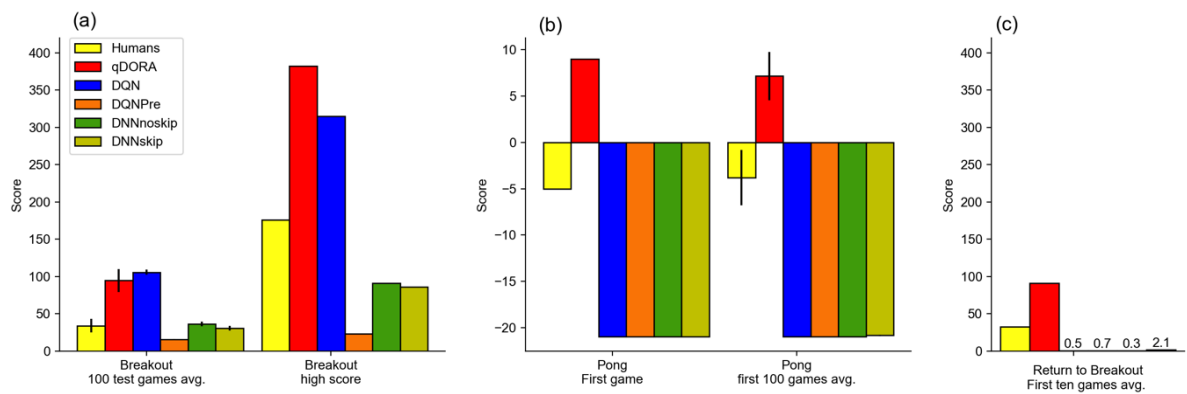

**Fig. 6.**
Results of game play simulations with Humans, DORA, the DQNs and the DNNs. Error bars represent 2 stderrors. (a) Performance humans and networks on Breakout as an average and high score. (b) Results of humans and networks playing Pong after training on Breakout (results are reported as reinforcement score calculated as own score minus opponent score). (c) Results of humans and networks when returning to play Breakout after playing or learning to play Pong.

---

[3] While we discuss the generalization in terms of learning to play Breakout and generalizing to playing Pong, the results are the same when training moves in the opposite direction (i.e., training on Pong and generalizing to Breakout).
[4] We use terms such as "moreY" or "higher" to describe the relations that DORA learned for the purposes of expositional clarity. For example, the predicate that we label "moreY" was the predicate representation that DORA learned that fit instances when one item was above another. DORA did not receive or utilize any such labels.

We then tested the capacity of the networks to play a new videogame, Pong. DORA had learned to play Breakout by learning associations between relational configurations and actions. Because DORA's representations of these relational configurations are effectively symbolic, they apply readily to untrained situations; that is, the predicates DORA uses to describe a relational configuration in Breakout can apply—or be bound to—items in any other situation (including novel situations; see also supplemental results). During its first game of Pong, DORA represented the game state using the relations it had learned playing Breakout. DORA discovered a correspondence between the action sets in the two games: particularly, *moreY/lessY* of the paddle (the paddle moves up and down) in Pong and *moreX/lessX* of the paddle (the paddle moves horizontally) in Breakout. This correspondence allowed DORA to infer via relational generalization (*16,20*; details in Methods) the relational configurations that reward specific moves in Pong. For example, just as *moreX* (ball, paddle) tends to reward a *moreX* move of the paddle in Breakout, *moreY* (ball, paddle) rewards a *moreY* move of the paddle in Pong.

Fig. 6B shows the performance of a human player and the networks on the first game of Pong after training on Breakout and the average performance over the first 100 games playing Pong. Like a human player, DORA performed at a high level on Pong on a single exposure to the game, and continued to play Pong at a high level. By contrast, all other networks showed poor performance—which is unsurprising given previous results using DNNs and transferring to different contexts (as noted above; *1-5*, *14-16*, *47*, *48*).

Finally, we trained the DQNs and DNNs on Pong. We then tested the capacity of the networks to return to playing Breakout. Fig. 6C shows the performance of a human player and the networks on the first game returning to play Breakout. Like a human player, DORA returned to playing the original game with little difficulty. By contrast, the other networks, showed poor performance, indicating that learning to play a new game interferes with the capacity to play old games.[5]

Importantly, the generalization failure of the networks using the same visual processing as DORA demonstrates that the visual processing does not produce representations that support the generalization performance demonstrated by DORA. Using both Q-learning and supervised learning, the unstructured (i.e., non-symbolic) representations produced by the visual processor and the DQN did not support extrapolatory generalization, while the predicate representations learned by DORA did.

**General Discussion**

We have shown that a machine system can perform extrapolatory generalization through *predicate learning*. DORA used predicate learning to discover symbolic representations from video game screen shots without feedback, and without assuming any structured representations a priori. Crucially, the predicate representations that DORA learned allowed it to extrapolate its knowledge to untrained situations. Specifically, the model was able to use the representations that it learned playing Breakout to successfully play a new game, Pong. Importantly, just as with human players, generalizing to a new game like Pong was fast (one-shot) and did not affect the system's ability to play the previously learned game. To our knowledge, we present the first demonstration of human-like generalization, or extrapolation, in a machine system that does not assume structured representations to begin with. Importantly, the solution makes use of well-established neurocomputational principles.

---

[5] While recent advances such as (*51*) have shown how such catastrophic forgetting can be avoided in neural networks, importantly, these systems rely on interleaved training (i.e., training on all to-be-learned tasks simultaneously). Block training, like we use (and which humans often engage in) still produces problems for DNNs.

Humans radically outperform machines in numerous domains, particularly when making inferences. The power of human generalization appears to stem from our ability to use structured representations (*e.g., 15-19*). DORA generalizes to untrained contexts because it uses structured representations to reason about those contexts. However, unlike other structured models, DORA learns the structured representations that it uses rather than assuming them a priori. The ability to discover structured representations from experience advances models of human cognition, exceeds the performance of current DNNs on video game transfer, and extends the principles of general machine learning by enabling extrapolatory generalization.

Of course, DORA still radically underperforms human behavior in general (humans do much more than play a few video games). People have much richer mental representations, which they acquire from a wide range of experiences in a myriad of contexts over their lives. However, the fact that we could approximate human-like performance using only structured representations of relative dimensions, speaks to the power of these representations (cf. *52-55*).

A predicate learning approach offers an account of how complex concepts might develop without the need to hardwire or encode the necessary information or structure into the system a priori. As such, we address limitations of current structure-based accounts of cognition (e.g., *15-19*), and offer a solution to the classic generalization problem that DNNs face (*1,3-5*). The ability to learn predicates from experience, and to generalize those representations to new inputs and contexts, may have substantial impact on theories of human reasoning, language representation and processing (*39,56*).

Though we do not assume innate representational types, we do not believe that the expressive power of our system negates the possibility that innate structures and biases exist in the human brain (*57*). Predicate learning makes a few clear claims about what infrastructure might be present in order to learn human-like structured representations. First, the system needs an innate or assumed ability to separate and compare streams of information (i.e., in banks of units connected to the same perceptual layer). Such a computational structure implies, for instance, a form of perceptual memory. Second, the system needs the capacity to learn to respond to temporal signals. In our network, this capacity is partially realized by layers of laterally inhibitive units. Beyond these computational primitives, other innate capacities or knowledge may also exist (e.g., biases to attend to specific kinds of information), but we do not need to assume them in order to learn representations that support human-like extrapolatory generalization.

The mechanism we have identified can be extended to more complex structures and to generalization across a wider range of experiences. This extension is tractable in two ways: first, under predicate learning, additional training, especially across different contexts and datasets, will result in richer representations which become more articulated and specified with experience. Second, predicate learning supports combining simple representations into more complex structures as needed (*58, 59*). The ability to identify useful combinations of simpler relational concepts can bootstrap learning more complex concepts. For instance, a complex concept like *support* might be built out of simpler concepts (e.g., a *contact* point of the supporting object *lower* than the contact point of the supported object). By combining our approach with more sophisticated reinforcement and model selection algorithms (e.g., *18,19*), we might find a strong way forward, with our system providing an account of where the structured representations that form the basis of more complex models come from in the first place.

Finally, we emphasize the fact that the model was able to discover and learn structured representations from experience because it exploits regularities of network behavior in a distributed computing system. We believe that this solution represents a fundamental formal and neurophysiological alignment between how human-like representations can be achieved in a system that learns, and how distributed neural computing systems, including cortical assemblies, process information (see *39* for a discussion; *24,25* for a historical precedent). Neural oscillations, or brain rhythms, (*6*) have long been implicated as indices of neural information processing (e.g., *60*) in many systems, even when those systems differ in scale and configuration (*61*) and have changed the nature of theories of the neural implementation of human memory (*62*) and of cortical speech and language processing (*63*). Learning symbolic structure from signals that naturally occur in distributed computing systems like the brain offers a promising avenue of research whereby the computational principles that yield the highest forms of the human mind (e.g., relational reasoning, formal and natural language processing) can be realized in systems that are based on the computational primitives of neurophysiological systems. Being able to generalize in a human-like way, or extrapolate symbolic structure, while using representations that are learned from experience is, in our view, a defining feature of the human mind. Capturing this ability in a machine system formalizes a core part of what it means to think like a human.

**Methods**

The current version of DORA is based on (*20*), with the addition of a local energy circuit that allows the model to learn invariant responses to instances of similarity and relative magnitude in pixel space. The DORA model is described in detail in (*20*) and also in the Supplemental Methods. Source code is available online (see Acknowledgements).

Below we describe the local energy circuit, and how it is learned. Throughout this section we use $n_i$ to denote input to unit $i$, $a_i$ to denote the activation of unit $i$, and $w_{ij}$ to denote the connection weight between units $i$ and $j$. We then describe methods for game play and generalization in DORA, and the other networks used for testing (see main text).

*Local energy circuit*

The local energy circuit consists of two layers of nodes. The top layer, $E$ takes input from any active predicate token units (we refer to predicate and object tokens as POs) in the focus of attention (FOA). The bottom layer, $A$ is connected to the feature units. Units in $E$ and $A$ are modified sigmoidal threshold units. Units in $E$ and $A$ are laterally inhibitive (i.e., units in $A$ inhibit all other units in $A$, and units in $E$ inhibit all other units in $E$).

Input, $n_i$, to units $i$ in $E$ and $A$ is calculated as:

$$n_i = \sum_j a_j w_{ij} - \rho_i$$

(1)

Where, for $E$ units, $j$ are all units connected to $i$ including all active POs (token units for predicates and objects; *20* or Supplemental Methods for details) in the FOA and all units $j \neq i$ in $E$, and $\rho_i$ is the refraction of unit $i$, and for $A$ units $j$ are all units connected to unit $i$, including all units in $E$ and all units $j \neq i$ in $A$. Weights between POs and units in $E$ are initialized to values between .01 and 1, weights between units in $E$ are initialized to -1. Weights between units and $E$ and A are initialized to random numbers between 0 and 0.9, and weights between units in $A$ are initialized to -1

The refraction, $\rho_i$ is given by the equation:

$$\rho_i = \frac{1}{.1 + .0001 e^x}$$

(2)

Where, $x$ is the number of iterations since unit $i$ last fired.

Activation, $a_i$ of units in $E$ is calculated using the function:

$$\Delta a_i = \gamma \frac{1}{1 + e^{-k(n_i - \theta_A)}} - \eta a_i - LI$$

(3)

Where, $\theta_A$ is the threshold on unit in $E$ (set randomly to a value between .4 and .8), $k$ is a ramp parameter, dictating the rate of rise of the sigmoid (=10), $\gamma$ is a growth parameter, $\eta$ is a leak parameter, and $LI$ is the activation of the local inhibitor (a global refresh signal given when no active PO units are present in the FOA (see *20* and supplemental methods for details)).

Activation, $a_i$ of units in $A$ is calculated using the function:

$$a_i = \frac{1}{1 + e^{-k(n_i - \theta_A)}} - LI \qquad (4)$$

Where, $\theta_A$ is the threshold on threshold unit in $A$ (=.5), and $k$=10.

Connection weights between units in $A$ and a subset (10) of feature units are set to random values of between 0 and 1.

Connections between units in $E$ and $A$, and between are updated by the equation:

$$\Delta w_{ij} = a_i(a_j - w_{ij}) \qquad (5)$$

Where, $i$ and $j$ refer to units in $A$ and $E$ respectively for learning connections between units in $E$ and $A$, and $A$ and feature units respectively for learning connections between $A$ and feature units.

During magnitude comparison, PO units compete via lateral inhibition to become active (*20*). PO units are connected to semantic units indicating their absolute magnitude, with greater magnitudes encoded by larger numbers of units (see above). As a consequence, when the two PO units code different absolute magnitudes, the PO unit connected to the greater magnitude will win the completion to become active and inhibit the PO unit connected to the lesser magnitude.

When the PO units settle, some $E$ units will fire most strongly when there is a single active PO unit (as when two different magnitudes are compared), and others when there are two active PO units (as when two similar magnitudes are compared). Units in $E$ pass activation to units in $A$. Active units in $A$ then pass activation to any feature units to which they are connected. Feature units that are most strongly connected to units in $A$ that are active early in firing become the invariants for "more". The active PO unit learns a connection to the "more" semantics by Eq. 5. When the inhibitor on the active PO fires, the active PO unit is inhibited to inactivity, and the local inhibitor (LI) fires. The LI inhibits units in $A$, allowing units in A that are active later in firing to become active. Feature units that are most strongly connected to units in $A$ that are active later in firing become the invariants for "less". The active PO unit learns a connection to the "less" semantics (by Eq. 5).

When both PO units code for the same extent, they settle into a stable state of mutual activation. Two active PO units will activate gating node G. Node G inhibits nodes E and passes activation the nodes in A.

*Learning representations from screens*

We used DORA to simulate learning structured representations from screen shots from the game Breakout. This simulation aims to mirror what happens when a child (or adult) learns from experience in an unsupervised manner (without a teacher or guide). We then used simple Q learning to teach DORA how to play Breakout using its learned representations.[6]

---

[6] We describe the results in terms of DORA learning to play Breakout and generalizing to Pong, but results were the same when run in the other direction (i.e., train on Pong and generalize to Breakout).

We started with screenshots from Breakout, experienced during 250 games with random move selection. Each screen from each game was processed with the visual pre-processor that identified objects and returned the raw pixel values as features of those objects. When learning in the world, objects have several extraneous properties. To mirror this point, after visual pre-processing, each object was also attached to a set of 100 additional features selected randomly from a set of 10000 features. These additional features were included to act as noise, and to make learning more realistic. (Without these noise features, DORA learned exactly as described here, only more quickly.)

DORA learned from object representations in an unsupervised manner. On each learning trial, DORA selected one pair of objects from the current screenshot at random. DORA attempted to characterize any relations that existed between the objects using any relations it has previously learned (initially, it had learned nothing, and so nothing was returned). DORA selected a dimension at random and ran the two objects through the local energy circuit over that dimension. If the semantics returned matched anything in LTM (e.g., "more" and "less" "x"), then DORA used that representation from LTM to characterize the current objects. DORA then ran (or attempted to run) its retrieval, mapping, local energy, predication, multi-place relation learning, and refinement routines (see *20* and above), and stored the representations that it learned. We placed one constraint on DORA's retrieval algorithm such that more recently learned items were favored for retrieval. Specifically, with probability .6, DORA attempted to retrieve from the last 100 representations that it had learned. This constraint followed our assumption that items learned more recently are more salient and more likely to be available for retrieval.

*Q-learning for game play*

We used DORA's learned relations to describe each screen in terms of relations between the different objects and thus to form a representation of any given screen state. For a given screen, DORA used the representations it had previously learned to represent the relations between objects. During predicate learning, DORA had learned a representation of successive states as earlier and later in game play, so DORA used the representation of key objects on the current screen and an immediately previous screen. For any pair of objects, if DORA had learned a representation that characterized the relation between the two objects (in LTM and as measured by the energy circuit), DORA used that representation the characterize the objects.

The relations were then used to form a table of encountered relational states and used Q-learning (Watkins, 1989) to learn the approximate action-value function for Breakout. We used a rule length constraint of two relations per state, reflecting the simplicity of the game and the WM capacity exhibited by humans (Cowen, 2001). We trained DORA decreasing the learning rate linearly from 0.1 to 0.05 and the exploration rate linearly from 0.1 to 0.01 throughout the training session. We saved the version of the table that yielded the maximum score during the session.

*Generalization in DORA*

To generalize between games, DORA uses mapping and relational generalization (*16,20*). Broadly, DORA uses the move selection it has learned previously, maps this representation to the move selection for Pong, and then generalizes a rule for move selection in Pong using relational generalization; both are established algorithms used in models of relational reasoning. First, the representation of the moves available to DORA (sampled randomly during the first exposure to Pong) were placed in the focus of attention. The rules learned during game play—represented using the predicates DORA learned—were placed in the active memory. DORA attempted to map the representations in the focus of attention to those in the active memory, and then used relational generalization to infer rules for move selections based on the mappings it discovered. We note that the predicates were learned by DORA, the rules are associations between game states and moves that are learned via tabular q-leaning.

For example, the moves *moreY* (paddle, paddle2), *moreY* (paddle2, paddle), *sameY* (paddle, paddle2)— i.e., the moves available in the game of Pong—would be placed in the focus of attention. During reinforcement learning, DORA learned that relations between the ball and paddle and the trajectory of the ball predicted proper move selection. Specifically, DORA learned that the state *moreX* (ball, paddle) with ball in state *moreX* (ball2, ball1) (i.e., the ball was currently at a position of further along that x-axis than it had been in the previous screen) predicted moving right, i.e., *moreX* (paddle2, paddle), that the state *moreX* (paddle, ball) and *moreX* (ball1, ball2) predicted moving left, i.e., *moreX* (paddle, paddle2), that the state *sameX* (ball, paddle) predicted making no move, i.e., *sameX* (paddle, paddle2), and so on. These representations were placed in the active memory. DORA then attempted to map the representations in the focus of attention to those in the active memory. Because of the shared relational and object similarity, *moreY* (paddle, paddle2) in the focus of attention mapped to *moreX* (paddle, paddle2) in the active memory, *moreY* (paddle2, paddle) in the focus of attention mapped to *moreX* (paddle2, paddle) in the active memory, and *sameY* (paddle, paddle2) in the focus of attention mapped to *sameX* (paddle, paddle2) in the active memory. Through relational generalization, DORA then filled in the missing rule information for the Pong game states.

More precisely, after mapping, DORA placed a representation of a rule in the focus of attention, and the mapped move from Pong in the active memory. For example, DORA might place the rule *moreX* (ball, paddle) & *moreX* (ball2, ball1), then move = right, i.e., *moreX* (paddle2, paddle) in the focus of attention, and the mapped *moreY* (paddle, paddle2) in the active memory (Fig. 7A). During mapping, DORA mapped the representations *moreX*(paddle2, paddle1) to *moreY* (paddle2, paddle1), reflecting the discovery of the correspondence between the move set in Breakout and in Pong (Fig. 7B; mappings depicted as green double-arrowed lines). The relational generalization algorithm is a self-supervised learning algorithm (see *8,12*). During self-supervised learning, if no token units are active in the active memory to match active token units in the focus of attention, DORA will activate token units in the active memory that match active token units in the focus of attention. As detailed in Eq. A15, when a token unit *j* in the focus of attention is active, it will produce a global inhibitory signal to all active memory units to which it does not map. A uniform inhibition in the active memory signals DORA to activate a unit of the same type (i.e., P, RB, PO) in the active memory as the active token unit in the focus of attention. Therefore, as the representation of *moreX* (ball, paddle) becomes active in the focus of attention (Fig. 7C), the units for *moreX* and *lessX* and paddle, all map to items in the active memory, while the units for ball and the *moreX*+ball, *lessX*+paddle, and the relation *moreX* (ball, paddle) map to nothing in the active memory. In response to this signal, DORA recruits nodes in the active memory with no strong connections to match the active and unmapped P, RB, and PO units in the focus of attention (Fig. 7D). These units then learn connections via Hebbian learning (Fig. 7E). The same process will occur as the representation of moreX (ball2, ball1) becomes active in the focus of attention (Fig. 7F-H). Through relational generalization, DORA matches the relational pattern of the focus of attention rule with the available units in the active memory. The result is a representation of structurally similar rule: *moreY* (ball, paddle) & *moreY* (ball2, ball1), move = up, i.e., *moreY* (paddle2, paddle) (Fig. 7I). This process was repeated for the other learned rules.

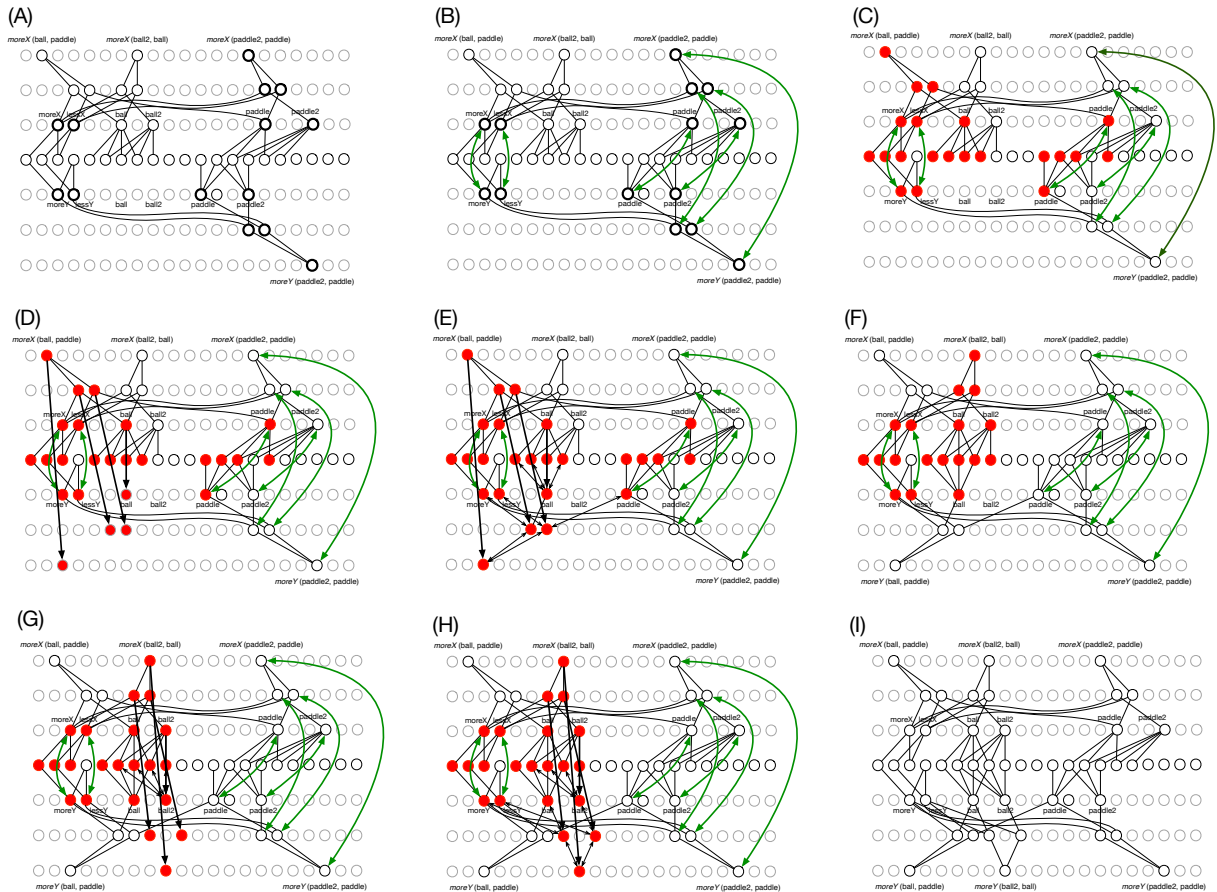

**Fig. 7.**
Graphical depiction of relational generalization in DORA. (A-B) a representation of the rule *moreX* (ball, paddle) & *moreX* (ball2, ball1), move = up, i.e., *moreX* (paddle2, paddle) is placed in the focus of attention, and mapped to the representation of *moreY* (paddle2, paddle1) in the active memory (green double arrowed lines). (C) The representation of *moreX* (ball, paddle) becomes active in the focus of attention, and some active units have nothing to map to in the active memory. (D) DORA recruits and activates units to match the unmapped focus of attention units, and learns connections between active token units in the active memory (open arrowed lines). The result is a representation of *moreY* (ball, paddle) in the active memory. (E-G) The same process occurs when *moreX* (paddle2, paddle1) becomes active in the focus of attention. (I) The end result is a representation of the rule: *moreY* (ball, paddle) & *moreY* (ball2, ball1), move = up, i.e., *moreY* (paddle2, paddle).

*Deep Q-Network*

A Deep Q-Network (DQN; *9*) was trained to play Breakout and Pong. The raw 210 × 160 frames were pre-processed by first converting their RGB representation to grey-scale and down-sampling it to a 105 × 80 image. We stacked the last 4 consecutive frames to form the input each state.

The input to the neural network was the 105 × 80 × 4 pre-processed state. The first hidden convolutional layer applied 16 filters of size 8 x 8 with stride 4 with a relu activation function. The second hidden convolutional layer applied 32 filters of size 4 x 4 with stride 2 with a relu activation function. The third hidden layer was fully connected of size 256 with a relu activation function. The output layer was fully connected with size 6 and a linear activation function.

We implemented all the procedures of (*9*) to improve training stability, in particular: (a) We used memory replay of size 1,000,000. (b) We used a target network which was updated every 10,000 learning iterations. (c) We fixed all positive rewards to be 1 and all negative rewards to be −1, leaving 0 rewards unchanged. (d) We clipped the error term for the update through the Huber loss.

We also ran the same network using the input from the visual preprocessor described above.

*Deep neural network*

We trained a deep neural network (DNN; *8*) in a supervised manner to play Breakout and Pong and tested generalization between games. One network was trained using random frame skipping and the other with fixed frame skipping.

The inputs to the network were the output of the visual preprocessor described above. Specifically, the network took as input the x and y positions of the ball and player-controlled paddle, as well as the left paddle for Pong (left as zeros when playing Breakout). The input to the neural network was a vector of size 24 corresponding to the pre-processed last seen 4 frames. This was fed to three fully connected layers of size 100 each with a relu activation function. The output layer was fully connected with size 6 and a softmax activation function.

The criteria for training was the correct action to take in order to keep the agent-controlled paddle aligned with the ball. In Breakout if the ball was to the left of the paddle the correct action las 'LEFT', if the ball was to the right of the paddle the correct action was 'RIGHT' and if the ball and the paddle were at the same level on the x-axis the correct action was 'NOOP'. In Pong if the ball was higher than paddle the correct action was 'RIGHT', if the ball was lower than paddle the correct action was 'LEFT' and if the ball and the paddle were at the same level on the y-axis the correct action was 'NOOP'. This action was encoded as a one-hot vector (i.e., activation of 1 for the correct action and cero for all other actions).

**Data and source code availability:**
The data generated and presented in the current study, and source code are available from github.com/alexdoumas/BrPong_18.

**Acknowledgments:**
This research was supported by grant ES/K009095/1 from Economic and Social Research Council of the United Kingdom to AEM. We thank Mante S. Nieuwland and Hugh Rabagliati for comments on an earlier version of this work, and Dylan Opdam and Zina Al-Jibouri for assistance with the initial research. Source code is available from github.com/alexdoumas/BrPong_18. Ethics and consent for human participants from Ethics Committee of the faculty of social science of Radboud University. Author contributions: LAAD and AEM designed research; LAAD and GP conducted research; LAAD and AEM wrote the manuscript. The authors declare no conflicts of interest.

**Supplemental results**

Additional tests of the symbolic adequacy of the learned representations

The representations that DORA learns should behave like structured relational representations. That is, the representations DORA learns must meet the requirements of human relational representations. While almost any structured representations will support analogical mapping some more substantive hallmarks of relational representations (see *22*, *73*) are that they, (i) form the basis for solving cross mappings; (ii) support mapping similar, but non-identical predicates; (iii) support mapping objects with no featural overlap, including completely novel objects, if they play similar roles; and (iv) form the basis of overcoming the n-ary restriction.

During a cross-mapping, an object (object1) is mapped to a featurally less similar object (object2) rather than a featurally more similar object (object3) because it (object1) plays the same role as the less similar object (object2). For example, if cat1 chases mouse1 and mouse2 chases cat2, then the structural analogical mapping places cat1 into correspondence with mouse2 because both play the *chaser* role. The ability to find such a mapping is a key property of genuinely relational (i.e., as opposed to feature-based) processing (see, e.g., *73-76*). Cross-mappings serve as a stringent test of the structure sensitivity of a representation as they require violating featural or statistical similarity.

We tested the relations that DORA had learned in the previous part of this simulation for their ability to support cross-mappings. We selected two of the refined relations that DORA had learned previously at random, such that both selected representations coded for the same relation (e.g., both coded for *taller*, or both coded for *same-width*). We bound the relations to new objects, creating two new propositions, P1 and P2 such that the agent of P1 was semantically identical to the patient of P2 and patient of P1 was semantically identical to the agent of P2. For example, P1 might be *taller* (square, circle) and P2 might be *taller* (circle, square). DORA then attempted to map P1 onto P2. We were interested in whether DORA would map the square in P1 onto the circle in P2 (the correct relational mapping) or simply map the square to the square and the circle to the circle. We repeated this procedure 10 times (each time with a different randomly-chosen pair of relations). In each simulation, DORA successfully mapped the square in P1 to the circle in P2 and vice-versa (because of their bindings to mapped relational roles). DORA's success indicates that the relations it learned in the first part of this simulation satisfy the requirement of supporting cross-mapping. DORA successfully solves cross-mappings because the correspondences that emerge between matching predicates and their corresponding RBs, during asynchronous binding force relationally similar objects into correspondence. For example, consider a case when DORA attempts to map *taller* (square, circle) in the focus of attention, and *taller* (circle, square) in the active memory. When the *more-height*+square role-binding becomes active in the focus of attention, because of asynchronous binding, the units coding for *more-height* will become active first, followed by the units coding for square. When *more-height* is active in the focus of attention, it will activate *more-height* and its corresponding RB, *more-height*+circle, in the active memory. When the units coding for square subsequently become active in the focus of attention, the active *more-height*+circle RB unit in the active memory (already in correspondence with the active *more-height*+square RB unit in the focus of attention) will activate the square unit, thus putting circle and square into correspondence, and allowing DORA to map them.

We then tested whether the relations that DORA had learned would support mapping to similar but non-identical relations (such as mapping *taller* to *greater-than*) and would support mapping objects with no semantic over-lap, including novel objects, that play similar roles. Humans successfully map such relations (*22*, *73*), an ability that depends on the semantic-richness of our relational representations. We selected two of the refined relations that DORA had learned during the previous part of this simulation, R1 and R2 (e.g, *taller*($x$,$y$) or *wider*($x$,$y$)). Crucially, R1 and R2 both coded for SRM across different dimensions (e.g., if R1 coded *taller*, then R2 coded *wider*). Thus, each role in R1 shared 50% of its semantics with a corresponding role in R2 (e.g., the role *more-height* has 50% of its semantics in common with the role *more-width*). To assure that no mappings would be based on object similarity, none of the objects that served as arguments of the relations had any semantic overlap at all. To ensure that the mapping would work with completely novel objects, we created objects composed from semantic units that we added to DORA solely for these simulations (i.e., these were semantic units DORA had not "experienced" previously). We repeated this process 10 times, each time with a different pair of relations from DORA's LTM. Each time, DORA mapped the agent role of R1 to the agent role of R2 and the patient role of R1 to the patient role of R2, and, despite their lack of semantic overlap, corresponding objects always mapped to one another (because of their bindings to mapped roles).

Finally, we tested whether the representations that DORA had learned would support violating the *n-ary restriction*: the restriction that an *n*-place predicate may not map to an *m*-place predicate when $n \neq m$. Almost all models of structured cognition follow the *n*-ary restriction (namely, those that represent propositions using traditional propositional notation and its isomorphs; see *15*). However, this limitation does not appear to apply to human reasoning, as evidenced by our ability to easily find correspondences between, say, *bigger* (Sam, Larry) on the one hand and *small* (Joyce) or *big* (Susan), on the other (*74*).

To test DORA's ability to violate the n-ary restriction, we randomly selected a refined relation, R1, that DORA had learned in the previous part of this simulation. We then created a single place predicate (r2) that shared 50% of its semantics with the agent role of R1 and none of its semantics with the patient role. The objects bound to the agent and patient role of R1 each shared 50% of their semantics with the object bound to r2. DORA attempted to map R1 to r2. We repeated this process 10 times, each time with a different relation from DORA's LTM, and each time DORA successfully mapped the agent role of R1 to r2, along with their arguments. We then repeated the simulation such that r2 shared half its semantic content with the patient (rather than agent) role of R1. In 10 additional simulations, DORA successfully mapped the patient role of R1 to r2 (along with their arguments). In short, in all our simulations DORA overcame the *n*-ary restriction, mapping the single-place predicate r2 onto the most similar relational role of the multi-place relation R1.

Finally, as noted above, DORA also learned representations of *greater* (*x,y*) and *same* (*x,y*) that were independent of any particular dimensions (i.e., relations that coded strongly for only the invariant features of *more* & *less* and *same*, that were otherwise not strongly connected to any other semantic features). Importantly, these representations also met all the requirements of structured relational representations. We ran the exact same tests for cross-mapping, mapping arguments with no semantic overlap based on shared roles, and violating the n-ary restriction that are described above, but using the representations of abstract magnitude (i.e., *greater* (*x,y*)) that DORA had learned during training. Just as with DORA's dimensional SRM representations, DORA's more abstract SRM representations successfully performed a cross-mapping in 10 out of 10 simulations, mapped arguments with no semantic overlap based only on shared roles in 10 out of 10 simulations and overcame the n-ary restriction in 10 out of 10 simulations.

**Supplemental methods: Details of DORA's Operation**

DORA's (see *20*) processing architecture is a layered network of bidirectionally connected units. The *long-term-memory* (LTM) bank consists of 3 layers of units (see Fig. 2). The bottom layers of LTM is connected to the feature (or semantic) layer. Additionally, DORA has 3 mutually exclusive banks of units, the *focus of attention/driver* (FOA), the *active memory/recipients* (AM), and the *emerging-active-memory/emerging-recipient* (EM). These banks are identical in structure to LTM, and also all connect to the feature unit bank. FOA controls the flow of activation in DORA. Units in FOA pass activation to the feature units. Because the feature units are connected to all other banks, activation flows from FOA to units in the other banks. Currently, we are agnostic as to whether FOA, AM, and the EM are implemented as processing windows that recruit specific units in LTM (as they are currently implementes; e.g., 63), or independent banks of units.

Initially, representations in DORA are delivered by the front-end processor as objects coded as flat feature vectors (see Figure S1A). (In terms of cortical computation, feature nodes can be thought of as aggregate units, perceptual representations, or activation states over networks.)

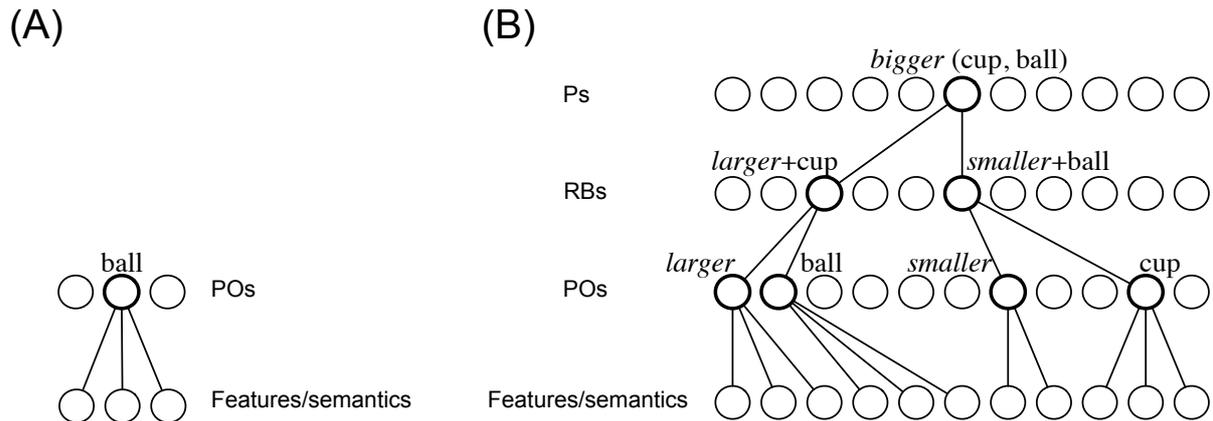

**Fig, S1.**
Representations in DORA. (a) DORA's starting state. DORA begins with representations of objects connected to features. (b) Representation of the proposition *bigger* (cup, ball) instantiated in layers of bidirectionally connected nodes. DORA learns propositional representations from examples of representations like those in (a).

DORA learns representations of a form we call LISAese (Figure S2B) via unsupervised learning (described in full below). Propositions in LISAese are coded by layers of units in a connectionist computing framework (Figure S2B). At the bottom of the hierarchy, feature (or semantic) nodes code for the featural properties of represented instances in a distributed manner. At the next layer, localist predicate and object units (POs) conjunctively code collections of semantic units into representations of objects and roles. At the next layer localist role-binding units (RBs) conjunctively bind object and role POs into linked role-filler pairs. Finally, proposition units (Ps) link RBs to form whole relational structures. Ps, RBs, and POs are our labels for the three layers of units in LTM, with POs the layer of units connected to the feature unit bank.

We use the term *analog* to refer to a complete story, event, or situation (e.g., from a single object in isolation, to a full propostion in LISAese). Analogs are represented by a collection of token units (P, RB and PO). Token units are not duplicated within an analog (e.g., within an analog, each proposition that refers to Don connects to the same "Don" unit). Separate analogs do have non-identical token units (e.g., Don will be represented by one PO unit in one analog and by a different PO in another analog). The feature units thus represent general type information and token units represent instantiations (or tokens) of those types in specific analogs (*15,16,20*).

In short, processing in DORA works as follows. Units in FOA become active (i.e., they are brought to DORA's attention). Units in FOA pass activation to the semantic units. Because the semantic units are shared by propositions in all sets, activation flows from FOA to propositions in the other three sets. All of DORA's operations (i.e., *retrieval*, *mapping*, *predicate learning*, *relation formation*, *schema induction*, and *generalization*, see below) proceed as a product of the units in FOA activating semantic units, which in turn activate units in the various other sets (as detailed below).

The model is based on a core set of neurocomputational computing principles. The model assumes (a) a layered architecture, with banks of layered units connected to a common pool of feature units. (b) Lateral inhibition between units in the layered banks. (c) Yoked inhibitors on units that accumulate input from their yoked units, and units at higher layers. (d) The capacity for Hebbian learning. There is ample evidence for all of these basic assumptions in the human neuroscience literature (*64*).

*Sequence of events in DORA*

The general sequence of events in DORA's operation is outlined below. Fig. 3 also provides an outline of the general processing algorithm in DORA. The details of each steps, along with the relevant equations and parameter values, are provided in the subsections that follow. DORA is very robust to the values of the parameters (see *20*). Throughout the equations in this section, we will use the variable *a* to denote a unit's activation, *n* its (net) input, and $w_{ij}$ to denote the connection from unit *i* to unit *j*.

An analog, *D* (selected at random, or based on the current game screen) enters FOA. Network activations are initialized to 0, and the firing order, *C*, or propositions in *D* is selected (for the current simulations, this selection is random and inconsequential, however, see (*15*) for a detailed description of how a system like DORA can set its own firing order according to the constraints of pragmatic centrality and text coherence). DORA can perform SRM calculation, do retrieval from LTM, perform analogical mapping, and do comparison-based unsupervised learning. Currently, the order of operations of these routines is fixed and set to the order: retrieval, SRM calculation, mapping, learning.

Step 1: Do basic SRM calculation

Behaviour of the SRM circuit as well as how this circuit is learned are described in the Methods section. During SRM, PO units get input by the equation:

$$n_i = \frac{\sum_j a_j w_{ij}}{1 + num(j)} - \sum_k a_k ,$$

(S1)

where where *j* are semantic units connected to *i*, *num(j)* is the number of semantic units *j*, and *k* are other PO units in the FOA. Activation of PO units is calculated by the leaky integrator function in EQ. S8.

Step 2. DORA operations

Repeat the following until each RB in *D* has fired three times if mapping, or once, otherwise.

Step 2.1. Update mode of all P units in FOA and AM

P units in all propositions operate in one of three modes: Parent, child, and neutral (*15,16,20*). P mode is important for representing higher-order relations (e.g., *R1(x, R2(y, z))*; *14*). In the current simulations, P mode did not need to change. We include this information here for the purposes of completeness.

Step 2.2. Update input to all token units in FOA

Token units in the FOA update their input by the equation:

$$n_i = \sum_j a_j w_{ij} G - \sum_k a_k - s \sum_m 3 a_m - 10 I_i ,$$

(S2)

where *j* are all units above unit *i* (i.e., Ps for RBs, RBs for POs), *G* is a gain parameter attached to the weight between the RB and its POs (POs learned via DORA's comparison-based predication algorithm—and thus with mode=1—have *G*=2 and 1 otherwise), *k* is all units in the FOA in the same layer as *i* (for POs, *k* is only those POs not connected to the same RB as unit *i*, and any P units currently in child mode; but see item *m*, from Eq. S2), *m* are PO units that are connected to the same RB (or RBs) as *i*, and $I_i$ is the activation of the PO inhibitor yoked to *i*. When DORA is operating in binding-by-asynchrony mode, $s = 1$; when it is operating in binding-by-synchrony mode (i.e., like LISA), $s = 0$.

Step 2.3. Update input to the PO and RB inhibitors

Every RB and PO unit is yoked to an inhibitor unit *i*. Both RB and PO inhibitors integrate input over time as:

$$n_i^{(t+1)} = n_i^{(t)} + \sum_j a_j w_{ij},$$

(S3)

where *t* refers to the current iteration, *j* is the RB or PO unit yoked to inhibitor unit *i*, and $w_{ij}$ is the weight between RB or PO inhibitor *i* and its yoked RB or PO unit (set to 1). Inhibitor units become active ($a_i = 1$) when $n_i$ is greater than the activation threshold (=220). RB inhibitors are yoked only to their corresponding RB. PO inhibitors are yoked both to their corresponding PO and all RB units in the same analog. As a result, at any given instant, PO inhibitors receive twice as much input as RB inhibitors, and reach their activation threshold twice as fast. POs, therefore, oscillate twice as fast as RBs. PO and RB inhibitors establish the time-sharing that carries role-filler binding information and allows DORA to dynamically bind roles to fillers. All PO and RB inhibitors become refreshed ($a_i = 0$ and $n_i = 0$) when the global inhibitor ($\Gamma_G$; described below) fires.

Step 2.4. Update the local and global inhibitors

The local and global inhibitors, $\Gamma_L$ and $\Gamma_G$ respectively (see e.g., *65-71*), serve to allow units in AM to keep pace with firing of units in FOA. The local inhibitor is inhibited to inactivity ($\Gamma_L = 0$) by any PO in FOA with activation above $\Theta_L$ (= 0.5), and becomes active ($\Gamma_L = 10$) when no PO in FOA has an activity above $\Theta_L$. During phase-lag-1 binding, the predicate and object POs time-share. There is a period during the firing of each role-filler pair after the one PO fires and before the other PO becomes active when no PO in FOA is very active. During this time the local inhibitor becomes active and inhibits all PO units in AM to inactivity. Effectively, $\Gamma_L$ serves as a local refresh signal, punctuating the change from predicate to object or object to predicate firing in FOA, and allowing the units in AM to keep pace with units in FOA.

The global inhibitor works similarly. It is inhibited to inactivity ($\Gamma_G = 0$) by any RB in FOA with activation above $\Theta_G$ (= 0.5) and becomes active ($\Gamma_G = 10$) when no RB in FOA is active above threshold. During the transition between RBs in FOA there is a brief period when no FOA RB are active above $\Theta_G$. During this time $\Gamma_G$ inhibits all units in AM to inactivity, allowing units in AM to keep pace with units in FOA.

Step 2.5. Update input to semantic units

Semantic units update their input as:

$$n_i = \sum_{j \in S \in (D,R)} a_j w_{ij},$$
(S4)

where $j$ is all PO units in $S$, which is the set of propositions in FOA, $D$, and AM $R$, and $w_{ij}$ is the weight between PO unit $j$ and semantic unit $i$.

Step 2.6. Update input to token units in AM and EM

Input to all token units in AM and emergent AM are not updated for the first 5 iterations after the global or local inhibitor fires.

All token units in AM and EM update their input by the equation:

$$n_i = \sum_j a_j w_{ij} + SEM_i + M_i - \sum_k a_k - s \sum_m 3a_m - \sum_n a_n - \Gamma_G - \Gamma_L,$$
(S5)

where $j$ is are any units above token unit $i$ (i.e., P units for RBs, RB units for POs; input from $j$ is only included on phase sets beyond the first), $SEM_i$ is the semantic input to unit $i$ if unit $i$ is a PO, and 0 otherwise, $M_i$ is the mapping input to unit $i$, $k$ is all units in either AM (if unit $i$ is in AM) or EM (if unit $i$ is in EM) in the same layer as $i$ (for POs, $k$ is only those POs not connected to the same RB as unit $i$, and any P units currently in child mode; but see item $m$, from Eq. S5), $m$ is PO units connected to the same RB (or RBs) as $i$ (or 0 for non-PO tokens), and $n$ is units above unit $i$ to which unit $i$ is not connected. When DORA is operating in binding-by-asynchrony mode, $s = 1$; when it is operating in binding-by-synchrony mode (i.e., like LISA), $s = 0$. $SEM_i$, the semantic input to $i$, is calculated as:

$$SEM_i = \frac{\sum_j a_j w_{ij}}{1 + num(j)},$$
(S6)

where $j$ are semantic units, $w_{ij}$ is the weight between semantic unit $j$ and PO unit $i$, and $num(j)$ is the total number of semantic units $i$ is connected to with a weight above $\theta$ (=0.1) (see 15,16,50). $M_i$ is the mapping input to $i$:

$$M_i = \sum_j a_j \left(3w_{ij} - Max(Map(i)) - Max(Map(j))\right),$$
(S7)

where $j$ are token units of the same type as $i$ in FOA (e.g., if $i$ is a RB unit, $j$ is all RB units in FOA), $Max(Map(i))$ is the highest of all unit $i$'s mapping connections, and $Max(Map(j))$ is the highest of all unit $j$'s mapping connections. As a result of Eq. S7, an active token unit in FOA will excite any AM unit to which it maps and inhibit all AM units of the same type to which it does not map.

Step 2.7. Update activations of all units in the network

All token units in DORA update their activation by the leaky integrator function:

$$\Delta a_i = \gamma n_i (1.1 - a_i) - \delta a_i]_1^0, \tag{S8}$$

where $\Delta a_i$ is the change in activation of unit $i$, $\gamma$ (=0.3) is a growth parameter, $n_i$ is the net input to unit $i$, and $\delta$ (=0.1) is a decay parameter. Semantic units update their activation by the equation:

$$a_i = \frac{n_i}{\max(n_i)}, \tag{S9}$$

where $a_i$ is the activation of semantic unit $i$, $n_i$ is the net input to semantic unit $i$, and $\max(n_i)$ is the maximum input to any semantic unit. There is physiological evidence for divisive normalization in the feline visual system (e.g., *42-44*) and psychophysical evidence for divisive normalization in human vision (e.g., *45,46*).

RB and PO inhibitors, $i$, update their activations according to a threshold function:

$$a_i = \begin{cases} 1, & n_i > \Theta_{IN} \\ 0, & otherwise \end{cases}, \tag{S10}$$

where $\Theta_{IN} = 220$.

Step 2.8. Update all mapping hypotheses

During the mapping process, DORA learns mapping hypotheses between all token units in FOA and token units of the same type in AM (i.e., between P units, between RB units and between PO units in the same mode [described below]). Mapping hypotheses initialize to zero at the beginning of a phase set. The mapping hypothesis between a FOA unit and an AM unit of the same type is updated by the equation:

$$\Delta h_{ij}^t = a_i^t a_j^t, \tag{S11}$$

where $a_i^t$ is the activation of FOA unit $i$ at time $t$.

Step 2.9 Run retrieval

DORA uses a variant of the retrieval routine described by (*15*).

Step 2.10. Run comparison-based unsupervised learning

DORA's comparison-based learning (CBL) routines are unsupervised. In the current version of the model, learning is licensed whenever 70% of FOA token units map to AM items (this 70% criterion is arbitrary, and in practice 100% of the units nearly always map). If learning is licensed DORA runs unsupervised learning. If FOA contains single objects, not yet bound to any predicates (i.e., each RB in FOA is bound only to a single PO), then DORA runs comparison-based predication learning. Otherwise, DORA runs refinement learning.

Step 2.10.1. Predicate learning

During comparison-based predicate learning (CBP) DORA attempts to recruit token units to learn conjunctive representations of units at lower layers. When only POs are active in AM, DORA recruits a new PO unit (i.e., a PO connected to no semantic features) in AM. The mode of the recruited PO is set to 1.

During CBP, DORA also attempts to recruit units in token layers above layers with active token units (e.g., in the RB layer if POs are active, or the P layer if RBs are active). As a result of lateral inhibition, if any tokens are already active in a layer, then the recruited unit cannot not become active. However, if no other token unit are active in a layer, the recruited unit can become active. The activation of any active recruited units is clamped to 1.o and remains at 1 until $\Gamma_G$ or $\Gamma_L$ fires. Connections between the new PO and all active semantics by the equation:

$$\Delta w_{ij} = a_i(a_j - w_{ij})\gamma \tag{S12}$$

where $\Delta w_{ij}$ is the change in weight between the new PO unit, $i$, and semantic unit, $j$, $a_i$ and $a_j$ are the activations of $i$ and $j$, respectively, and $\gamma$ is a growth rate parameter. Connections between corresponding active token units (i.e., between P and RB, or RB and PO units) are also updated by Eq. S12, where $I$ is the recruited unit, and $j$ a a token units in a adjacent layers. In this present code, the process of recruiting units during CBL is accomplished by fiat, but several neutrally-plausible solutions to the problem of unit recruitment have been proposed, including those described in previous DORA papers (e.g., *20*; see also *72*).

Thus, DORA exploits the temporal pattern that emerges when sets corresponding objects are compared to learn representations of single-place predicates, and the temporal pattern that emerges when sets of corresponding of role-filler pairs are mapped in order to link RBs into multi-place relations. When the phase set ends, connection weights between the new P unit $i$ and any RBs to which it has connections, $j$, are updated by the equation:

$$w_{ij} = \begin{cases} 1, & w_{ij} > 0 \text{ and } \sum_k w_{ik} \geq 2 \\ 0, & \text{otherwise} \end{cases}, \tag{S13}$$

for P unit $i$ and RB unit $j$, and $k$ is all RB units (including $j$) in AM.

Step 2.10.2. Refinement Learning

*Step 2.10.2.1: Predicate refinement.* For any PO in FOA that is currently active, and maps to a unit in AM with a mapping connection above the threshold $\Theta_{MAP}$ (=0.7), DORA infers a PO unit connected to no semantic features in the EM with a mapping connection to the active FOA unit. DORA learns connections between the new PO and all active semantics by Eq. S12. DORA also licenses self-supervised learning (SSL). During SSL, DORA infers token units in the EM that match active tokens in $D$ (FOA). Specifically, DORA infers a structure unit in the EM in response to any unmapped token unit in $D$. If unit $j$ in $D$ maps to nothing in the EM, then when $j$ fires, it will send a global inhibitory signal to all units in the EM (Eq. S7). This uniform inhibition, unaccompanied by any excitation in AM us a signal that DORA exploits, and infers a unit of the same type (i.e., P, RB, PO) in the EM. Inferred PO units in the EM have the same mode as the active PO in FOA. The activation of each inferred unit in the EM is set to 1. DORA learns connections between corresponding active tokens in the EM (i.e., between P and RB units. and between RB and PO units) by Eq. S12 (where unit *j* is the newly inferred token unit, and unit *i* is any other active token unit). To keep DORA's representations manageable (and decrease the runtime of the simulations), at the end of the phase set, we discard any connections between semantic units and POs whose weights are less than 0.1.

*Step 2.10.2.2: Relational generalisation.* The relational generalisation algorithm is a self-supervised learning algorithm adopted from that used in (*16*). During self-supervised learning, if no token units are active in AM to match active token units in FOA, DORA will activate token units in AM that match active token units in FOA. As detailed in Eq. S7, when a token unit *j* in FOA is active, it will produce a global inhibitory signal to all AM units to which it does not map. A uniform inhibition in AM signals DORA to activate a unit of the same type (i.e., P, RB, PO) in AM as the active token unit in FOA. DORA learns connections between corresponding active tokens in the EM (i.e., between P and RB units. and between RB and PO units) by the simple Hebbian learning rule in Eq. S12 (where unit *j* is the newly active token unit, and unit *i* is the other active token unit). Connections between PO units and semantic units are updated by Eq. S12.

Step 3. Update mapping connections

Mapping connections are updated at the end of each phase set. First, all mapping hypotheses are normalized by the equation:

$$h_{ij} = \left(\frac{h_{ij}}{MAX(h_i, h_j)}\right) - MAX(h_{kl}) , \qquad (S14)$$

where, $h_{ij}$ is the mapping hypothesis between units *i* and *j*, $MAX(h_i, h_j)$ is the largest hypothesis involving either unit *i* or unit *j*, and $MAX(h_{kl})$ is the largest mapping hypothesis where either k=*i* and l≠*j*, or l=*j* and k≠*i*. That is, each mapping hypothesis is normalised divisively: Each mapping hypothesis, $h_{ij}$ between units *i* and *j*, is divided by the largest hypothesis involving either unit *i* or *j*. Next each mapping hypothesis is normalized subtractively: The value of the largest hypothesis involving either *i* or *j* (not including $h_{ij}$ itself) is subtracted from $h_{ij}$. The divisive normalization keeps the mapping hypotheses bounded between zero and one, and the subtractive normalization implements the one-to-one mapping constraint by forcing mapping hypotheses involving the same *i* or *j* to compete with one another (see *7*). Finally, the mapping weights between each unit in FOA and the token units in AM of the same type are updated by the equation:

$$\Delta w_{ij} = \eta(1.1 - w_{ij})h_{ij}]_1^0 , \qquad (S15)$$

where $\Delta w_{ij}$ is the change in the mapping connection weight between FOA unit *i* and AM unit *j*, $h_{ij}$ is the mapping hypothesis between unit *i* and unit *j*, $\eta$ is a growth parameter, and $\Delta w_{ij}$ is truncated for values below 0 and above 1. After each phase set, mapping hypotheses are reset to 0. The mapping process continues for three phase sets.

**Supplemental results**

Additional tests of the symbolic adequacy of the learned representations

The representations that DORA learns should behave like structured relational representations. That is, the representations DORA learns must meet the requirements of human relational representations. While almost any structured representations will support analogical mapping some more substantive hallmarks of relational representations (see *22, 73*) are that they, (i) form the basis for solving cross mappings; (ii) support mapping similar, but non-identical predicates; (iii) support mapping objects with no featural overlap, including completely novel objects, if they play similar roles; and (iv) form the basis of overcoming the n-ary restriction.

During a cross-mapping, an object (object1) is mapped to a featurally less similar object (object2) rather than a featurally more similar object (object3) because it (object1) plays the same role as the less similar object (object2). For example, if cat1 chases mouse1 and mouse2 chases cat2, then the structural analogical mapping places cat1 into correspondence with mouse2 because both play the *chaser* role. The ability to find such a mapping is a key property of genuinely relational (i.e., as opposed to feature-based) processing (see, e.g., *73-76*). Cross-mappings serve as a stringent test of the structure sensitivity of a representation as they require violating featural or statistical similarity.

We tested the relations that DORA had learned in the previous part of this simulation for their ability to support cross-mappings. We selected two of the refined relations that DORA had learned previously at random, such that both selected representations coded for the same relation (e.g., both coded for *taller*, or both coded for *same-width*). We bound the relations to new objects, creating two new propositions, P1 and P2 such that the agent of P1 was semantically identical to the patient of P2 and patient of P1 was semantically identical to the agent of P2. For example, P1 might be *taller* (square, circle) and P2 might be *taller* (circle, square). DORA then attempted to map P1 onto P2. We were interested in whether DORA would map the square in P1 onto the circle in P2 (the correct relational mapping) or simply map the square to the square and the circle to the circle. We repeated this procedure 10 times (each time with a different randomly-chosen pair of relations). In each simulation, DORA successfully mapped the square in P1 to the circle in P2 and vice-versa (because of their bindings to mapped relational roles). DORA's success indicates that the relations it learned in the first part of this simulation satisfy the requirement of supporting cross-mapping. DORA successfully solves cross-mappings because the correspondences that emerge between matching predicates and their corresponding RBs, during asynchronous binding force relationally similar objects into correspondence. For example, consider a case when DORA attempts to map *taller* (square, circle) in the focus of attention, and *taller* (circle, square) in the active memory. When the *more-height*+square role-binding becomes active in the focus of attention, because of asynchronous binding, the units coding for *more-height* will become active first, followed by the units coding for square. When *more-height* is active in the focus of attention, it will activate *more-height* and its corresponding RB, *more-height*+circle, in the active memory. When the units coding for square subsequently become active in the focus of attention, the active *more-height*+circle RB unit in the active memory (already in correspondence with the active *more-height*+square RB unit in the focus of attention) will activate the square unit, thus putting circle and square into correspondence, and allowing DORA to map them.

We then tested whether the relations that DORA had learned would support mapping to similar but non-identical relations (such as mapping *taller* to *greater-than*) and would support mapping objects with no semantic over-lap, including novel objects, that play similar roles. Humans successfully map such relations (*22, 73*), an ability that depends on the semantic-richness of our relational representations. We selected two of the refined relations that DORA had learned during the previous part of this simulation, R1 and R2 (e.g, *taller*($x,y$) or *wider*($x,y$)). Crucially, R1 and R2 both coded for SRM across different dimensions (e.g., if R1 coded *taller*, then R2 coded *wider*). Thus, each role in R1 shared 50% of its semantics with a corresponding role in R2 (e.g., the role *more-height* has 50% of its semantics in common with the role *more-width*). To assure that no mappings would be based on object similarity, none of the objects that served as arguments of the relations had any semantic overlap at all. To ensure that the mapping would work with completely novel objects, we created objects composed from semantic units that we added to DORA solely for these simulations (i.e., these were semantic units DORA had not "experienced" previously). We repeated this process 10 times, each time with a different pair of relations from DORA's LTM. Each time, DORA mapped the agent role of R1 to the agent role of R2 and the patient role of R1 to the patient role of R2, and, despite their lack of semantic overlap, corresponding objects always mapped to one another (because of their bindings to mapped roles).

Finally, we tested whether the representations that DORA had learned would support violating the *n-ary restriction*: the restriction that an *n*-place predicate may not map to an *m*-place predicate when $n \neq m$. Almost all models of structured cognition follow the *n*-ary restriction (namely, those that represent propositions using traditional propositional notation and its isomorphs; see *15*). However, this limitation does not appear to apply to human reasoning, as evidenced by our ability to easily find correspondences between, say, *bigger* (Sam, Larry) on the one hand and *small* (Joyce) or *big* (Susan), on the other (*74*).

To test DORA's ability to violate the n-ary restriction, we randomly selected a refined relation, R1, that DORA had learned in the previous part of this simulation. We then created a single place predicate (r2) that shared 50% of its semantics with the agent role of R1 and none of its semantics with the patient role. The objects bound to the agent and patient role of R1 each shared 50% of their semantics with the object bound to r2. DORA attempted to map R1 to r2. We repeated this process 10 times, each time with a different relation from DORA's LTM, and each time DORA successfully mapped the agent role of R1 to r2, along with their arguments. We then repeated the simulation such that r2 shared half its semantic content with the patient (rather than agent) role of R1. In 10 additional simulations, DORA successfully mapped the patient role of R1 to r2 (along with their arguments). In short, in all our simulations DORA overcame the *n*-ary restriction, mapping the single-place predicate r2 onto the most similar relational role of the multi-place relation R1.

Finally, as noted above, DORA also learned representations of *greater* (*x,y*) and *same* (*x,y*) that were independent of any particular dimensions (i.e., relations that coded strongly for only the invariant features of *more & less* and *same*, that were otherwise not strongly connected to any other semantic features). Importantly, these representations also met all the requirements of structured relational representations. We ran the exact same tests for cross-mapping, mapping arguments with no semantic overlap based on shared roles, and violating the n-ary restriction that are described above, but using the representations of abstract magnitude (i.e., *greater* (*x,y*)) that DORA had learned during training. Just as with DORA's dimensional SRM representations, DORA's more abstract SRM representations successfully performed a cross-mapping in 10 out of 10 simulations, mapped arguments with no semantic overlap based only on shared roles in 10 out of 10 simulations and overcame the n-ary restriction in 10 out of 10 simulations.